%% file: root.tex
\documentclass[letterpaper, 10 pt, conference]{ieeeconf}  

\IEEEoverridecommandlockouts                              

\usepackage{graphicx} 
\usepackage{amsmath} 
\usepackage{amssymb}  
\usepackage{amsfonts}
\usepackage{color}
\usepackage{empheq}
\usepackage{algorithm}
\usepackage{algorithmic}
\usepackage{ulem}

\title{\LARGE \bf Active Inference for Autonomous Decision-Making with \\ Contextual Multi-Armed Bandits}

\author{Shohei Wakayama and Nisar Ahmed$^*$
\thanks{$^*$Work supported by the NASA COLDTech Program, grant \#80NSSC21K1031. S. Wakayama was also supported by the Masason Foundation. The authors are with the Smead Aerospace Engineering Sciences Department, University of Colorado Boulder, Boulder, CO 80303 USA        {\tt\small [shohei.wakayama; nisar.ahmed]@colorado.edu}}
}

\begin{document}

\maketitle
\thispagestyle{empty}
\pagestyle{empty}

\begin{abstract}
In autonomous robotic decision-making under uncertainty, the tradeoff between exploitation and exploration of available options must be considered. If secondary information associated with options can be utilized, such decision-making problems can often be formulated as contextual multi-armed bandits (CMABs). In this study, we apply active inference, which has been actively studied in the field of neuroscience in recent years, as an alternative action selection strategy for CMABs. Unlike conventional action selection strategies, it is possible to rigorously evaluate the uncertainty of each option when calculating the expected free energy (EFE) associated with the decision agent's probabilistic model, as derived from the free-energy principle. We specifically address the case where a categorical observation likelihood function is used, such that EFE values are analytically intractable. We introduce new approximation methods for computing the EFE based on variational and Laplace approximations. Extensive simulation study results demonstrate that, compared to other strategies, active inference generally requires far fewer iterations to identify optimal options and generally achieves superior cumulative regret, for relatively low extra computational cost. 
\end{abstract}

\section{INTRODUCTION} \label{sec: introduction}
\input{introduction.tex}

\section{BACKGROUND and RELATED WORK} \label{sec: background}
\input{background.tex}

\section{METHODOLOGIES} \label{sec: methodologies}
\input{methodologies}

\section{SIMULATION STUDY} \label{sec: simulation_study}
\input{simulation_study}

\section{CONCLUSIONS} \label{sec: conclusions}
\input{conclusion.tex}
\bibliographystyle{IEEEtran}
\bibliography{root.bib}

\end{document}

%% file: introduction.tex
In robotics, autonomous robots must often optimize choices from among multiple alternatives to accomplish a task under uncertainty. 
Such decision-making can be complex when the outcomes 
obtained for different options are stochastic and their distributions are unknown {\it a priori}. If a robot could devote unlimited time and resources to the decision-making process, it would be possible to simulate all options a significant number of times and derive accurate outcome distributions before finalizing the best option. 
However, in many environments where autonomous robots are actually expected to be deployed, time and resource (e.g. bandwidth and energy) 
constraints are severe. This can also hamper the ability of human users to effectively interact with robots. In such settings, lightweight and effective algorithms are needed for decision-making with limited information. 
\begin{figure}[t]
    \centering
    \includegraphics[width=5.5cm]{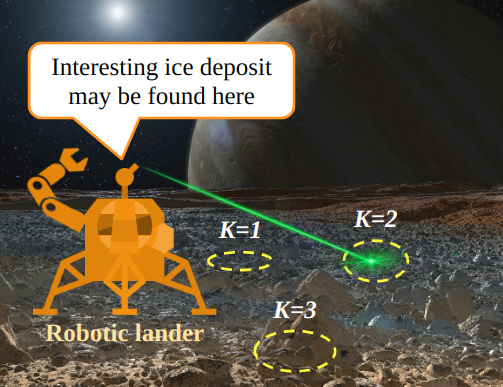}
    \caption{A robotic lander reasons about whether candidate dig sites on an icy moon contain scientifically interesting data. To speed up the process, it is desirable not only to balance exploitation and exploration, but also incorporate a selection bias, i.e. prior preferences, for desired observations.}
    \label{fig: example_scenario}
    \vspace{-0.2in}
\end{figure}

For instance, as illustrated in Fig. \ref{fig: example_scenario}, suppose a robot in a remote environment (such as an icy solar system moon) is assigned a task of collecting scientifically interesting objects (e.g. ice deposits and minerals) from a single search area using a resource-intensive manipulator. Although the robot could send image and spectral data about each area to human experts and let them decide where to investigate, mission time and communication constraints (e.g. due to high radiation and hours long round trip time delays for speed of light signals) make smooth human-robot interaction hard in such missions \cite{pappalardo2014, jay2022}. Thus, the robot must autonomously and accurately infer the outcome probability that each area contains valuable objects, while using sensors that are easily deployable and consume little energy. To improve efficiency, the robot must strike a balance between increasing the certainty of plausible areas ({\it exploitation}) and reducing the uncertainty of unknown areas ({\it exploration}), taking into account the surrounding environment, mission status, etc. 

Such a decision-making problem can be formulated as a contextual multi-bandit (CMAB), a type of reinforcement learning problem which has originally been studied and applied in recommendation systems \cite{li2010} and recently utilized in robotics \cite{steve2019, chan2019}. 
Generally speaking, however, bandit problems (depending on their scale and complexity) typically require a large number of iterative interactions with the problem environment to finalize an optimal option,
which can be a bottleneck in applying them to many practical robotic applications like space exploration. 
Therefore, to address such concerns, this paper considers {\it active inference} \cite{friston2015, kaplan2018} as an action selection strategy for CMAB problems. Active inference has recently attracted attention in the field of neuroscience because it can mathematically rigorously take into account the uncertainty of each option in the process by calculating a measure named {\it expected free energy (EFE)} and incorporate a selection bias toward preferred outcomes. In this way, it is theoretically possible to determine the best option with fewer iterations. In fact, active inference has been applied to standard stationary and dynamic switching MABs and those properties have been experimentally verified \cite{markovic2021}, but no theory or studies have yet applied it to the important case of contextual bandits. Also, for CMABs with discrete categorical action outcomes, the EFE terms are generally not tractable for general context-dependent likelihood models. Thus, we propose two novel statistical EFE approximation methods based on variational inference and Laplace approximation, to permit the use of active inference for solving CMABs. These methods are demonstrated and validated in a simulated autonomous search area selection problem, and their performances are compared with existing state of the art CMAB action selection strategies. 
%
%
For the remainder of the paper:
Sec. \ref{sec: background} reviews the MAB and CMAB problems, as well as an overview of active inference and the free-energy principle behind the theory; Sec. \ref{sec: methodologies} formulates the action selection strategy for CMAB problems using active inference, and derives the EFE approximations; Sec. \ref{sec: simulation_study} presents the setup and results of the simulation experiment (motivated by the autonomous remote exploration example); and Sec. \ref{sec: conclusions} presents conclusions.


%% file: background.tex
\subsection{Multi Armed Bandits (MABs) and Contextual MABs} \label{sec: cmab}
The multi-armed bandit (MAB) is a classic reinforcement learning problem of optimal selection from multiple alternatives with uncertain outcomes. The MAB allows a decision-making agent like an autonomous robot to account for the tradeoff between exploitation (i.e. preferring alternatives with the higher expected outcome) and exploration (i.e. preferring alternatives with a more uncertain distribution of outcomes) \cite{kuleshov2014}. Although seemingly simpler than other sequential decision-making problems such as the Markov decision process (MDP) \cite{bellman1957, thrun2005} and the partially observable Markov decision process (POMDP) \cite{kaelbling1998, kurniawati2022}, the theory of MABs has long been the subject of research in the field of statistical machine learning and has been applied to solving problems in the fields of recommendation systems, healthcare, and finance \cite{bouneffouf2020}. While there are several major formulations of MABs such as stochastic bandits \cite{auer2002} and adversarial bandits \cite{auer1995}, depending on how uncertain outcomes are handled, here we focus on stochastic bandits treating both present and future outcomes in the same fashion, i.e. without any discounting. This simplifies the development and is appropriate in settings where it may not be possible to predict in advance how many iterations will be required for an agent to learn an action selection policy.  


More formally written, a stochastic MAB problem aims to minimize the cumulative regret, i.e. the sum of difference between the maximum expected outcome and an actual outcome 
over iterations, via alternately performing two steps: (1) action selection: selecting one option (a.k.a ``arm") from alternatives with unknown outcome distributions; and (2) observation update: updating the statistics for the selected option based on the outcome obtained. 
In stochastic bandits, it is common to apply the Bayes' rule to update the posterior of the estimated outcome distribution of the selected option in step (2). For step (1), various action selection methods such as $\epsilon$-greedy, the upper confidence bound (UCB) \cite{auer2002}, and the Thompson sampling (TS) \cite{thompson1933} have been devised. 


In standard MAB problems, only the information about each option itself (e.g. the cumulative value of outcomes and/or the number of times it is selected) is taken into account during action selection. However, it may be possible to minimize the cumulative regret more effectively and find the best option with fewer iterations, by utilizing contextual information associated with each option (e.g. user age in case of web advertisement). Variants of stochastic MABs that incorporate such secondary information are called \emph{contextual multi-armed bandits (CMABs)} and have been used for dialogue systems, anomaly detection, etc. \cite{bouneffouf2020}. In robotic decision-making problems, it is typical to collect and use multiple sources of side information, so we will focus on CMAB problems hereafter. 
{However, conventional action selection strategies for CMABs use heuristics (e.g. the exploration bonus term in UCB) to select options, and these methods usually require many iterations for informed decision-making \cite{li2010, kuleshov2014}. Thus, it is desirable to develop an alternative action selection strategy for robotics applications where the number of iterations cannot be too large. }
\subsection{Free-Energy Principle and Active Inference} \label{sec: free_energy_active_inference}

The free-energy principle is a theory in neuroscience that is expected to explain whole-brain mechanisms \cite{friston2010}. According to this principle, biological agents perceive, learn, and act by minimizing free-energy, which is the upper bound of surprise (i.e. $-\log p(o)$, where $o$ is an observation/outcome) observed from the external world, to maintain their homeostasis according to neurally encoded probabilistic models. In neuroscience, this principle has been used to explain many physiological phenomena, such as predictive coding \cite{friston2009}.   

Active inference is a mathematical framework that applies the free-energy principle to the behavioral norms of biological agents, and has recently begun to attract attention in not only neuroscience and but also robotics \cite{lanillos2021}. Since future observations are unobservable until an action is executed, in active inference, decision-making agents are assumed to act to minimize a measure called expected free energy (EFE). The reason why this framework is theoretically sound for sequential decision-making problems is that the EFE is composed of value and information gain terms. Thus, by minimizing the EFE, it is possible to naturally balance the tradeoff between exploitation and exploration in a principled non ad-hoc manner.  Additionally, active inference can easily incorporate prior information about the probability distribution of outcomes that agents desire to observe, which is called a \emph{prior preference} (a.k.a ``evolutionary prior"). 
By adjusting this distribution, agent behavior can be varied. 
{Adjustment of the prior preference values corresponds to a type of reward tuning required in typical sequential decision-making problems. Yet, specifying the probability of observed outcomes desired by the agent arguably provides a theoretically intuitive and elegant alternative to specifying numerical reward values over intermediate actions and states.} \footnote{There exist studies attempting to estimate an appropriate prior preference in terms of reward functions via inverse reinforcement learning \cite{shin2022}, however this type of learning is not the focus of our study.} 


However, active inference has only recently been considered for multi-armed bandit problems, namely in \cite{markovic2021}. Their work has shown that an active inference action selection strategy produces higher performance (i.e. smaller cumulative regrets) than the state-of-the-art methods, especially for stationary MABs when the number of total iterations are relatively small and for dynamic switching MABs. Nevertheless, their study does not focus on contextual bandits, which are arguably even more applicable and interesting for autonomous robotics. Thus, in the following, after reviewing CMAB problems (where in particular outcomes are categorical), we explain how the EFE expression is derived in these problems. We then derive novel EFE approximations to address the fact that the resulting terms are analytically intractable due to the observation likelihoods. 


%% file: methodologies.tex
\subsection{Problem Statement} \label{sec: problem_statement}

In a contextual multi-armed (CMAB) problem introduced in Section \ref{sec: cmab}, suppose the total number of options (actions $a$) taken into account by an autonomous robot is $K$. These options correspond to the bandit arms. 
Also suppose that the observed outcome $o_k$ of each option/action $a$ is binary, i.e. $o_k \in \left\{0,1\right\} \ \forall k \in \left\{ 1,\cdots,K \right\}$.  
\footnote{The methodology described in this section is extendable to more than two outcome categories with a few straightforward modifications.} 
Note, we use the shorthand the notation $a_k \leftrightarrow a=k$. 

Since the number of outcomes is binary, the probability $\psi_k$ that each option $k$ outputs a preferred outcome, $o_k=1$, depends on a hidden linear parameter vector $\vec{\theta}_k$ (unique to each option) and a shared context vector $\vec{x}$, and is represented by the following sigmoid function (without loss of generality, $\vec{x}$ is assumed common for all options and iterations), 
\begin{eqnarray} \label{eq: sigmoid}
    \psi_k = p(o_k = 1) = \frac{e^{\vec{\theta}_k^T \vec{x}}}{1+e^{\vec{\theta}_k^T \vec{x}}} = \frac{1}{1 + e^{- \vec{\theta}_k^T \vec{x}}}.
\end{eqnarray}
In the case of such a Bernoulli contextual bandit, the expected outcome of each option is also $\psi_k$ so that the cumulative regret is expressed as follows,   
\begin{eqnarray} \label{eq: regret}
    \mbox{Regret}(T) = T\psi^* - \sum_{k=1}^K N_{T}(k)\psi_k,
\end{eqnarray}
where $T$ is the total number of iterations, $\psi^*$ is the maximum expected outcome (unknown {\it a priori} for the robot), and $N_{T}(k)$ represents how many times an option $k$ is selected within $T$ iterations. The goal of CMAB problems is to minimize expected cumulative regret. To do so, the robot needs to estimate the values of the hidden parameter vectors $\vec{\theta}_k \ \forall k, \ k \in \{1, \cdots, K \}$ in the process of finding an optimal option by iteratively performing the two steps of action selection and measurement update. 



\subsection{Expected Free Energy in CMABs} 
As already mentioned in Sec. \ref{sec: free_energy_active_inference}, according to the free-energy principle, an agent 
wants to avoid (i.e. minimize) surprise to maintain homeostasis. The surprise in the case of the CMAB problem defined in Sec. \ref{sec: problem_statement} can be expressed by marginalizing the following joint distribution,
{\small 
\begin{align} \label{eq: surprise}
    -\log p(o) = -\log \int_{\vec{\theta}_1} \cdots \int_{\vec{\theta}_K} p(o, \vec{\theta}_1, \cdots, \vec{\theta}_K) d\vec{\theta}_1 \cdots d\vec{\theta}_K. 
\end{align}}However, calculating (\ref{eq: surprise}) directly via multiple integral is often analytically intractable, so instead its upper bound (i.e. free-energy) is minimized. Yet, outcomes are unobservable until an option is executed, in the case of active inference, the expected free energy (EFE) is considered. Note, in the following, $\vec{\Theta}$ represents $[\vec{\theta}_1, \cdots, \vec{\theta}_K]^T$, 
{\small 
\begin{align} \label{eq: efe_cmab_general}
    &\mbox{EFE}(a_k) = \int_{\vec{\Theta}}  q(\vec{\Theta}|a_k) \sum_o p(o|\vec{\Theta}, a_k) \log \frac{q(\vec{\Theta}|a_k)}{p(\vec{\Theta}, o|a_k)} d\vec{\Theta}.
\end{align}}where $q(\vec{\Theta}|a_k)$ is a proposal distribution that approximates the posterior distribution $p(\vec{\Theta}|o, a_k)$ and $p(o|\vec{\Theta}, a_k)$ is an observation likelihood. By further transforming the log term in (\ref{eq: efe_cmab_general}) as follows
{\small
\begin{eqnarray}
    \log \frac{q(\vec{\Theta}|a_k)}{p(\vec{\Theta},o|a_k)} &=& \log \frac{q(\vec{\Theta}|a_k)}{p(\vec{\Theta}|o, a_k)p(o|a_k)} \nonumber \\
    &\approx& \log \frac{q(\vec{\Theta}|a_k)}{p(\vec{\Theta}|o, a_k)} + \log \frac{1}{p_{ev}(o)} \nonumber \\
    &=& \log \frac{q(\vec{\Theta}|a_k)p(o, a_k)}{p(\vec{\Theta}, o, a_k)} + \log \frac{1}{p_{ev}(o)} \nonumber \\
    &=& \log \frac{q(\vec{\Theta}|a_k)p(o|a_k)p(a_k)}{p(o|\vec{\Theta}, a_k)p(\vec{\Theta}|a_k)p(a_k)} + \log \frac{1}{p_{ev}(o)} \nonumber \\
    &\approx& \log \frac{q(o|a_k)}{p(o|\vec{\Theta}, a_k)} + \log \frac{1}{p_{ev}(o)} \label{eq: approximated_log_term}
\end{eqnarray}
}and substituting (\ref{eq: approximated_log_term}) back to (\ref{eq: efe_cmab_general}), the EFE is approximated as 
{\small
\begin{align}
    \mbox{EFE}(a_k) \approx \int_{\Theta} q(\vec{\Theta}|a_k) \sum_o p(o|\vec{\Theta}, a_k) \log \frac{q(o|a_k)}{p_{ev}(o)\cdot p(o|\vec{\Theta}, a_k)} d\vec{\Theta}, \label{eq: efe_cmab_general_approximated}
\end{align}
}where $q(o|a_k)$ is the marginalization of $q(o,\vec{\Theta}|a_k)$ and $p_{ev}(o)$ is a prior preference. 
For brevity, by further assuming that the hidden linear parameter vectors are independent each other, eq.(\ref{eq: efe_cmab_general_approximated}) can be rewritten as 
{\footnotesize
\begin{align}
    &\mbox{EFE}(a_k) \approx \nonumber \\ 
    &\int_{\vec{\theta}_k} q(\vec{\theta}_k|a_k) \sum_o p(o|\vec{\theta}_k, a_k) \Big\{\log \frac{q(o|a_k)}{p_{ev}(o)} - \log p(o|\vec{\theta}_k, a_k) \Big\} d\vec{\theta}_k \nonumber \\
    &= -\sum_o \int_{\vec{\theta}_k} q(\vec{\theta}_k|a_k)p(o|\vec{\theta}_k, a_k) d\vec{\theta}_k \cdot \log p_{ev} (o) \nonumber \\
    &\hspace{0.13in} -\sum_o \int_{\vec{\theta}_k} q(\vec{\theta}_k|a_k)p(o|\vec{\theta}_k, a_k) \cdot \log \frac{p(\vec{\theta}_k|o, a_k)}{p(\vec{\theta}_k|a_k)} d\vec{\theta}_k \nonumber \\
    &= -\sum_o q(o|a_k) \cdot \log p_{ev}(o) \nonumber \\
    &\hspace{0.13in} -\sum_{o} q(o|a_k) \int_{\vec{\theta}_k} q(\vec{\theta}_k|o,a_k) \cdot \log \frac{p(\vec{\theta}_k|o,a_k)}{p(\vec{\theta}_k|a_k)} d\vec{\theta}_k \nonumber \\
    &= -\mathbb{E}_{q(o|a_k)} \Big[\log p_{ev}(o) \Big] - \mathbb{E}_{q(o|a_k)} \Big[D_{KL}(q(\vec{\theta}_k|o, a_k)||q(\vec{\theta}_k|a_k))\Big]. \label{eq: efe_each_option}
\end{align}}As can be seen in (\ref{eq: efe_each_option}), minimizing EFE value naturally achieves balancing exploitation (the first term) and exploration (the second term). 
Thus, in CMAB problems with active inference, the robot needs to choose the option minimizing the EFE term 
in each action selection step, 
\begin{equation} \label{eq: action_selection}
    a_{selected} = \underset{a \in |A|}{\arg \min} \ \mbox{EFE} (a).
\end{equation}
Yet, as the observation likelihood $p(o|\vec{\theta}_k, a_k)$ is the sigmoid function, the EFE term cannot be computed analytically. Hence, in the following subsections, we introduce the two statistical approximation methods: (1) variational Bayesian importance sampling (VBIS) \cite{nisarTRO}, which has been applied to other probabilistic decision-making under uncertainty problems in robotics \cite{lukeTRO, wakayamaTRO}; and (2) Laplace approximation \cite{bishop2006}, which is commonly used in statistical machine learning. 
{Note that, instead of the sigmoid function, a conditional probability table (CPT) \cite{bishop2006} could be used to express the observation likelihood and achieve tractable posterior expressions for the EFE (and CMAB inference more generally using other approaches like $\epsilon$-greedy). However, this is generally far less practical, since the number of CPT parameters increases exponentially with the dimension of $\vec{x}$, whereas this only increases linearly for the sigmoid likelihood.}

\subsection{VBIS EFE Approximation} \label{sec: vbis}
The variational Bayesian importance sampling (VBIS) is a hybrid data fusion method that approximates an analytically intractable posterior distribution in human-robot collaborative sensing when a flexible discrete-continuous function, e.g. softmax function, is used as an observation likelihood \cite{nisarTRO}.
The main idea behind the VBIS algorithm is that a softmax function can be lower-bounded by a variational Gaussian via the inequality proved in \cite{bouchard2007} so that if a prior distribution is a Gaussian, a posterior distribution is also approximated as another Gaussian. Here, since the logistic sigmoid function in the EFE expression is a special case of a softmax function with two classes, the following log-sum-exponential (LSE) function is upper bounded as follows
{\small
\begin{align} \label{eq: bouchard_inequality}
\begin{split} 
    \log \Big(\sum_{h=0}^1 e^{y_h} \Big) &\leq \alpha + \sum_{h=0}^1 \frac{y_h - \alpha - \xi_h}{2}  \\ &+\lambda(\xi_h)\Big[(y_h - \alpha)^2 - \xi_h^2 \Big] + \log(1+e^{\xi_h}),
\end{split}
\end{align}}\noindent where $y_0 = 0$, $y_1 = \vec{\theta}^T \vec{x}$, $\lambda(\xi_h) = \frac{1}{2\xi_h}[\frac{1}{1+e^{-\xi_h}} - \frac{1}{2}]$. $\alpha$ and $\xi_h$ are free variational parameters which can be adjusted to achieve the tighter bound of the LSE and be iteratively updated within the VBIS procedure. By replacing the denominator of (\ref{eq: sigmoid}) with this bound, the sigmoid function is lower bounded by the variational Gaussian. 
\begin{eqnarray}
    p(o=j|\vec{\theta}_k) \geq \mbox{exp} \Big( \mathcal{G} + \mathcal{H}^T \vec{\theta}_k - \frac{1}{2} \vec{\theta}_k^T \mathcal{K} \vec{\theta}_k \Big), 
\end{eqnarray}
where $\mathcal{G}$, $\mathcal{H}$, and $\mathcal{K}$ are
\begin{equation}
    \mathcal{G} = \sum_{h=0}^1 \Big[ \frac{\xi_h}{2} + \lambda(\xi_h)(\xi_h^2 - \alpha^2) - \log(1+e^{\xi_h}) \Big], 
\end{equation}
\begin{empheq}[left={\mathcal{H}=\empheqlbrace}]{alignat=2}
  & -\frac{1}{2}\vec{x} + 2\lambda(\xi_1) \cdot \alpha \cdot \vec{x} &\qquad &\text{if $o=0$,} \\
  & \frac{1}{2}\vec{x} + 2\lambda(\xi_1) \cdot \alpha \cdot \vec{x}                 &       &\text{if $o=1$,}
\end{empheq}
\begin{equation}
    \mathcal{K} = 2\lambda(\xi_1) \cdot \vec{x} \cdot \vec{x}^T.
\end{equation}
Thus, in our problem, if we assume that the prior distribution of a hidden linear parameter vector $q(\vec{\theta}_k|a_k)$ is approximated as a (multivariate) Gaussian, the joint distribution $q(\vec{\theta}_k|a_k) \cdot p(o|\vec{\theta}_k, a_k)$ in (\ref{eq: efe_each_option}) becomes also another unnormalized Gaussian. By normalizing this distribution, the approximated normalization constant $\hat{C}_{vb} = \int_{\vec{\theta}_k} q(\vec{\theta}_k|a_k) \cdot \mbox{exp} \Big( \mathcal{G} + \mathcal{H}^T \vec{\theta}_k - \frac{1}{2} \vec{\theta}_k^T \mathcal{K} \vec{\theta}_k \Big) d\vec{\theta}_k$ required for calculating (\ref{eq: efe_each_option}) can be obtained from the well-known Gaussian integral. During this process, the VB posterior, which approximates the posterior $p(\vec{\theta}_k|o, a_k)$, is also derived as a side product. 
{\small
\begin{align}
    p(\vec{\theta}_k|o, a_k) &\approx  p_{vb}(\vec{\theta}_k|o, a_k) \nonumber \\ 
    &=  \frac{q(\vec{\theta}_k|a_k) \cdot \mbox{exp}(\mathcal{G} + \mathcal{H}^T \vec{\theta}_k - \frac{1}{2} \vec{\theta}_k^T \mathcal{K} \vec{\theta}_k)}{\int q(\vec{\theta}_k|a_k) \cdot \mbox{exp}(\mathcal{G} + \mathcal{H}^T \vec{\theta}_k - \frac{1}{2} \vec{\theta}_k^T \mathcal{K} \vec{\theta}_k) d\vec{\theta}_k}
\end{align}}\noindent However, the value of $\hat{C}_{vb}$ is smaller than the true normalization constant $C = \int_{\vec{\theta}_k} q(\vec{\theta}_k|a_k) p(o|\vec{\theta}_k, a_k) d\vec{\theta}_k$, the importance sampling (IS) is applied to improve the estimate by following \cite{nisarTRO}.

IS can approximate the expected value of an arbitrary function $f(\vec{\theta}_k)$ over a pdf such as a posterior that is difficult to sample directly, by utilizing a proposal distribution that is roughly similar in shape to the posterior and easy to sample. Here, in particular, the IS approximation is constructed as 
{\small
\begin{equation}
    \langle f(\vec{\theta}_k) \rangle = \langle \vec{\theta}_k \rangle \approx \sum_{s=1}^{N_s} w_s \vec{\theta}_{k,s}, w_s \propto \frac{q(\vec{\theta}_{k,s}|a_k)p(o|\vec{\theta}_{k,s}, a_k)}{p_{prop}(\vec{\theta}_{k,s})},
\end{equation}}where $N_s$ is the number of samples, $w_s$ is the importance weight for sample $s$, and $p_{prop}(\vec{\theta}_k)$ is the proposal pdf 
\begin{equation}
    p_{prop}(\vec{\theta}_k) = \mathcal{N}(\vec{\mu}_{vb}, \Sigma_{prior})
\end{equation}
where $\vec{\mu}_{vb}$ is the VB mean and $\Sigma_{prior}$ is the prior covariance matrix of $q(\vec{\theta}_k|a_k)$. During this procedure, the improved normalization constant estimate $\hat{C}_{vbis}$ can be calculated as $\sum_s {w_s}/N_s$. Thus, the first term inside the summation over outcome $o$ in the second equation of (\ref{eq: efe_each_option}) is rewritten as 
{\small
\begin{align}
    \log \frac{q(o|a_k)}{p_{ev}(o)} \cdot \int_{\vec{\theta}_k} q(\vec{\theta}_k|a_k) p(o|\vec{\theta}_k, a_k) d\vec{\theta}_k = \Big\{ \log \frac{C_{vbis}}{p_{ev}(o)} \Big\} \cdot C_{vbis}. \label{eq: efe_1st_final}
\end{align}}Similarly, the second term is approximated as 
{\small
\begin{align}
    &\int_{\vec{\theta}_k} q(\vec{\theta}_k|a_k) p(o|\vec{\theta}_k, a_k) \log p(o|\vec{\theta}_k, a_k) d\vec{\theta}_k \nonumber \\
    &= C_{vbis} \int_{\vec{\theta}_k} p_{vbis}(\vec{\theta}_k|o, a_k) \Big(\mathcal{G} + \mathcal{H}^T \vec{\theta}_k - \frac{1}{2} \vec{\theta}_k^T \mathcal{K} \vec{\theta}_k \Big) d\vec{\theta}_k. \label{eq: efe_2nd_term}
\end{align}
}However, the quadratic term in (\ref{eq: efe_2nd_term}) (i.e. $\mathcal{G} + \mathcal{H}^T \vec{\theta}_k - \frac{1}{2} \vec{\theta}_k^T \mathcal{K} \vec{\theta}_k$) has not yet been processed via the IS step. Therefore, it is required to re-update the lower-bounded variational Gaussian from the following identical equation to avoid underestimating an EFE value. 
{\small
\begin{equation} \label{eq: backup_sigmoid}
    C_{vbis}\cdot \mathcal{N}(\vec{\mu}_{vbis},\Sigma_{vbis}) = \mathcal{N}(\vec{\mu}_{prior}, \Sigma_{prior}) \cdot \mathcal{N}(\vec{\mu}_{sigm}, \Sigma_{sigm}) 
\end{equation}}After re-deriving $\mathcal{G}'$, $\mathcal{H}'$, and $\mathcal{K}'$ terms of $\mathcal{N}(\vec{\mu}_{sigm}, \Sigma_{sigm})$, (\ref{eq: efe_2nd_term}) is further transformed as follows. 
{\small
\begin{align}
    (\ref{eq: efe_2nd_term}) &= C_{vbis} \cdot \mathbb{E}_{p_{vbis}(\vec{\theta}_k|o,a_k)} \Big[ \mathcal{G}' + \mathcal{H}'^T \vec{\theta}_k - \frac{1}{2} \vec{\theta}_k^T \mathcal{K}' \vec{\theta}_k \Big] \nonumber \\
    &= C_{vbis}\Big(\mathcal{G}' + \mathcal{H}'^T \vec{\mu}_{vbis} - \frac{1}{2} \Big(Tr(\mathcal{K}'\Sigma_{vbis}) + \vec{\mu}_{vbis}^T \mathcal{K}' \vec{\mu}_{vbis} \Big)\Big) \label{eq: efe_2nd_final}
\end{align}}After subtracting (\ref{eq: efe_2nd_final}) from (\ref{eq: efe_1st_final}), and sum the EFE$(a_k, o)$ term over possible outcomes $o$, the EFE$(a_k)$ is computed. Then, by following (\ref{eq: action_selection}), an active inference robot finalizes the option in an action selection step. Algorithm \ref{alg: ai_cmab} summarizes the process of active inference action selection for CMABs with the VBIS EFE approximation.

\begin{algorithm}[t] 
\caption{Active inference action selection for CMABs} \label{alg: ai_cmab}
{\footnotesize
\begin{algorithmic}[1] 
\renewcommand{\algorithmicrequire}{\textbf{Input:}}
\renewcommand{\algorithmicensure}{\textbf{Output:}}
\REQUIRE Context vector $\vec{x}$, evolutionary prior $p_{ev}(o)$, estimated means and covariances, number of samples for importance sampling $N_s$  
\ENSURE a selected index for action selection 
\FOR {each option}
\FOR {each outcome}
\STATE obtain $\vec{\mu}_{posterior}$, $\Sigma_{posterior}$, and $\hat{C}_{posterior}$ from the VBIS or the Laplace approximation algorithm \cite{nisarTRO, bishop2006} 
\STATE re-derive $\mathcal{G}'$, $\mathcal{H}'$, $\mathcal{K}'$ terms from (\ref{eq: backup_sigmoid}) 
\STATE calculate $\mbox{EFE}(a_k, o)$ via (\ref{eq: efe_1st_final}) and  (\ref{eq: efe_2nd_final})  
\ENDFOR
\STATE calculate EFE$(a_k) = \sum_o \mbox{EFE}(a_k, o)$
\ENDFOR
\RETURN a selected index $ = \underset{a \in |A|}{\arg \min} \mbox{EFE}(a)$ 
\end{algorithmic} 
}
\end{algorithm} 

\subsection{Laplace EFE Approximation}
In statistical machine learning, the Laplace approximation is commonly used to approximate a continuous probability density function as a Gaussian distribution \cite{bishop2006}. This uses the second-order approximation of the vector Taylor expansion of a logarithmic function whose gradient is $\vec{0}$. Particularly, in the approximation of the EFE value in (\ref{eq: efe_each_option}), a function $g(\vec{\theta}_k)$ is defined as $q(\vec{\theta}_k|a_k)p(o|\vec{\theta}_k, a_k)$, and the following logarithmic function is used. 
\begin{align}
    \log g(\vec{\theta}_k) &\approx \log g(\vec{\theta}_{k}^{(0)}) + \sum_{r=1}^n (\theta_{k,r} - \theta_{k,r}^{(0)})\frac{\partial g(\vec{\theta}_k^{(0)})}{\partial \theta_{k,r}} \nonumber \\
    &+ \frac{1}{2} \Big\{\sum_{r=1}^n (\theta_{k,r} - \theta_{k,r}^{(0)}) \frac{\partial \log g(\vec{\theta}_{k}^{(0)})}{\partial \theta_{k,r}} \Big\}^2, \label{eq: second_order_approximation}
\end{align}
where $\vec{\theta}_k^{(0)}$ satisfies $\nabla \log g(\vec{\theta}_k^{(0)}) = \vec{0}$ and $n$ is the length of $\vec{\theta}_k$. However, since it is analytically intractable to find $\vec{\theta}_k^{(0)}$, $\vec{\theta}_{k,MAP}$ is computed via Newton's method \cite{galantai2000}. Also, the third term in (\ref{eq: second_order_approximation}) is rewritten as 
{\small
\begin{equation}
    \Big\{ \sum_{r=1}^n (\theta_{k,r} - \theta_{k,r}^{(0)}) \frac{\partial \log g(\vec{\theta}_{k}^{(0)})}{\partial \theta_{k,r}} \Big\}^2 = \Big\{ (\vec{\theta}_k - \vec{\theta}_k^{(0)})^T \nabla \log g(\vec{\theta}_k^{(0)}) \Big\}^2, 
\end{equation}}
therefore, (\ref{eq: second_order_approximation}) reduces to 
{\small
\begin{align}
    \log &g(\vec{\theta}_k) \approx \log g(\vec{\theta}_{k, MAP}) + \nonumber \\ &\frac{1}{2} (\vec{\theta}_k - \vec{\theta}_{k, MAP})^T H\Big[\log g(\vec{\theta}_{k, MAP})\Big] (\vec{\theta}_k - \vec{\theta}_{k, MAP}), \label{eq: laplace_with_hessian}
\end{align} 
}where $H$ is the Hessian. Thus, if we define $A = - H$ and by taking the logarithm from (\ref{eq: laplace_with_hessian}), 
{\small
\begin{equation}
    g(\vec{\theta}_k) \approx g(\vec{\theta}_{k,MAP}) \cdot \mbox{exp} \Big(-\frac{1}{2} (\vec{\theta}_k - \vec{\theta}_{k,MAP})^T A (\vec{\theta}_k - \vec{\theta}_{k,MAP})\Big), 
\end{equation}}and the normalization constant $C_{laplace}$ is computed as 
\begin{equation}
    C_{laplace} = g(\vec{\theta}_{k,MAP}) \cdot\frac{(2\pi)^{\frac{n}{2}}}{|A|^{\frac{1}{2}}}.
\end{equation}
After following the same procedures as with the VBIS described in Algorithm \ref{alg: ai_cmab} (i.e. calculating the Laplace posterior $\mathcal{N}(\vec{\mu}_{laplace}, \Sigma_{laplace})$ and re-deriving $\mathcal{G}'$, $\mathcal{H}'$, and $\mathcal{K}'$ terms of $\mathcal{N}(\vec{\mu}_{sigm}, \Sigma_{sigm})$), the EFE$(a_{k})$ of each option $k$ is calculated, and then the best option can be selected.  

%% file: simulation_study.tex


\subsection{Motivating Problem Scenario} 
NASA recently launched the \underline{C}oncepts for \underline{O}cean worlds \underline{L}ife \underline{D}etection \underline{Tech}nology (COLDTech) program to research and develop software for `fail-operational' autonomous robotic landers that will be dispatched to icy solar system moons like Europa and Enceladus \cite{jay2022}. Currently, the concept of operations calls for a robotic lander to perform basic surface tasks such as digging, imaging, moving objects, retrieving/analyzing samples with various instruments, and downlinked selected science data. 
As shown in Fig. \ref{fig: example_scenario}, a key capability will be autonomous selection of surface areas to prioritize for science investigation. Although human experts can specify preferences for possible areas in advance, the unknown dynamic nature of icy moon environments can cause science priorities to shift unexpectedly,
e.g. due to plume eruptions or changes in surface conditions. Such priorities must be considered in light of contextual information pertaining to overall mission progress, lander state, and engineering factors. 



In the following, we consider proof of concept CMAB simulations for a robotic lander which has been given the task of collecting an ice sample specimen from one of $K$ non-overlapping designated reachable areas within the landing site. The candidate sampling sites are selected by scientists based on imaging data collected immediately after landing. Since a sampling action consumes a considerable amount of resources (e.g. energy and time for grinding, drilling, scooping, etc.), the lander cannot just deploy a resource-intensive manipulator in a haphazard way. It is necessary to infer in advance which areas can yield scientifically valuable information, i.e. $o_k=1$, using a laser spectrometer similar to those used on Martian rovers \cite{manne2017determination}, which can be more easily and cheaply deployed than the manipulator tools. 
Panoramic images of the landing area taken onboard are also processed by a learning-based computer vision algorithm to produce a $C$-dimensional binary context feature vector $\vec{x}$, to encode various factors such as sun direction, surface reflectance, etc. \cite{matthies2007, palafox2017}. 
The autonomous site selection problem can thus be formulated as a CMAB, where the goal is to find the best site $k$ to sample with the manipulator, using observations $o_k=1/0$ (interesting/uninteresting science returns) from selected spectrometer target sites $a_k$ and context $\vec{x}$. 



\subsection{Simulation Setup and Results}
As described in Sec. \ref{sec: problem_statement}, CMAB problems require alternately executing two steps: action selection (which site to illuminate with laser spectrometer) and observation update (process spectrometer returns). In this simulation study, the following solution approaches are considered and compared in an extensive Monte Carlo study: (i) best possible action (site) selection, using an offline oracle (needed to compute regret); (ii) $\epsilon$-greedy (where $\epsilon = 0.25$ was found to work best after initial trials); (iii) upper confidence bound (UCB); (iv) logistic Thompson sampling (LTS); and (v) active inference (AI). The action selection methods for (iv) and (v) are paired with VBIS and Laplace approximations for the observation updates.
\begin{figure}[t]
    \centering
    \includegraphics[width=0.5\textwidth]{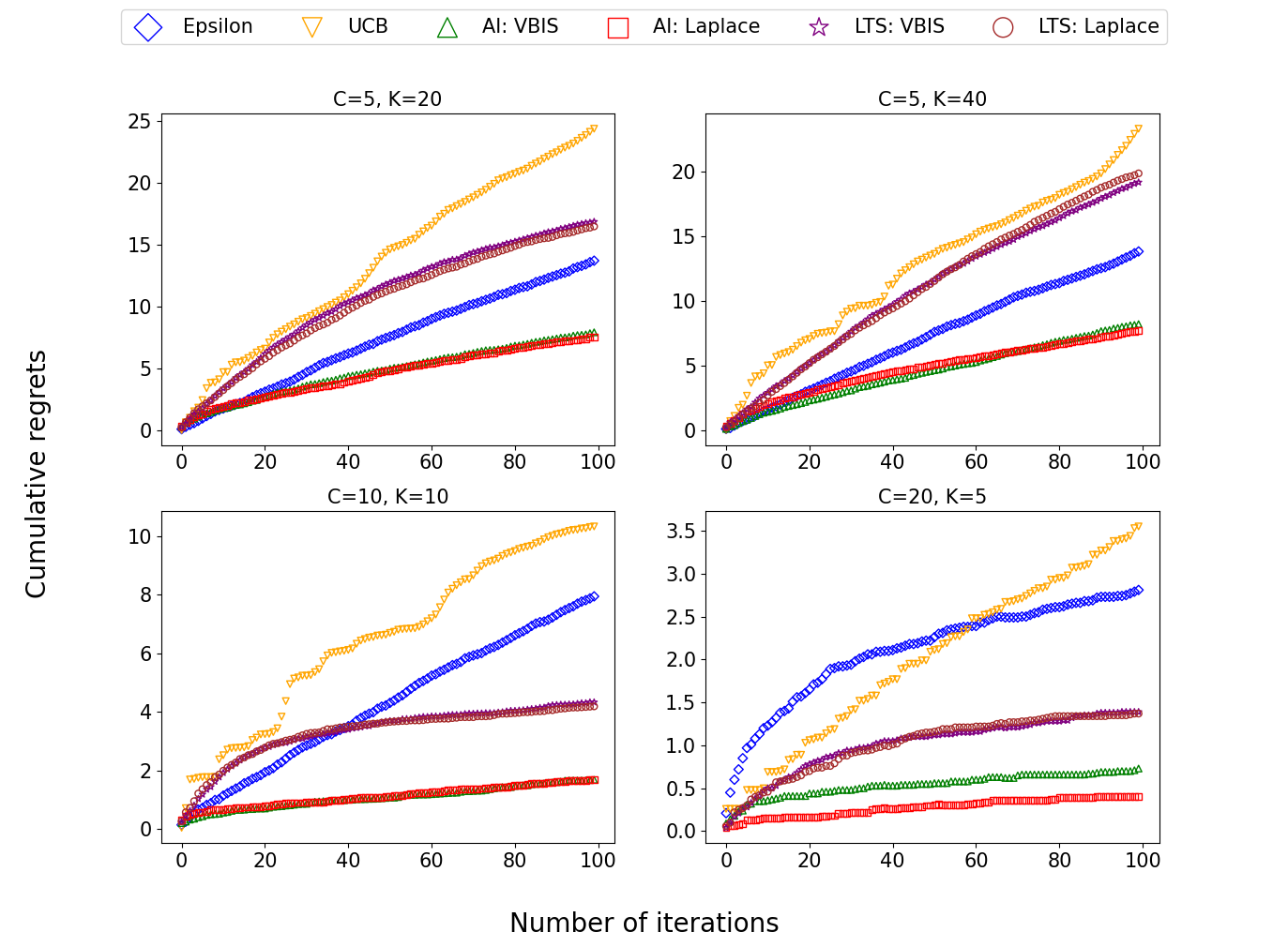}
    \caption{Comparison of cumulative regrets among different approaches. For each subplot, the cumulative regret is the average of $100$ simulations. Prior preference of active inference agents is set as $p_{ev}(o) = [0.001, 0.999]^T$. 
    }
    \label{fig: comparison_cumulative_regrets}
    \vspace{-0.15in}
\end{figure}
In all simulations, $100$ Monte Carlo (MC) runs are performed, and the number of iterations $T$ in each MC run is set to $10^2$, which is small compared to common MAB algorithm benchmarks \cite{markovic2021} and reflects a practical upper limit for robotic lander sensor deployment. Thus, there is no need to be too concerned about the suboptimal asymptotic behavior of active inference agents on very long time scales (e.g. $10^5$ iterations) as experimentally showed in \cite{markovic2021}. 
The true hidden linear parameters $\vec{\theta}_{k}$ for each candidate search site $k$ were randomly generated from a multivariate Gaussian distributions, 
{where the mean 
is sampled from a uniform distribution of $0$ to $1$, and the covariance matrix is generated using the result of 
a singular value decomposition on the Gram matrix of a matrix with uniformly randomly distributed entries}. 
The search context vector $\vec{x}$ was randomly generated assuming that each element takes a binary value with uniform probability, and the same value is used for all sites. 
To compare how each approach performs with respect to the difficulty of CMAB, the number of areas $K$ and search context elements $C$, 
and the evolutionary prior $p_{ev}(o)$, were varied, such that $K \in \{5, 10, 20, 40\}$, $C \in \{5, 10, 20\}$, and $p_{ev}(o) \in \{ [0.001, 0.999]^T, [0.4, 0.6]^T\}$, i.e. in total, $24 \times 7 = 168$ sets of 100 MC sims were run. 
This setup produces difficult CMABs since there can be multiple ``good" (large $\psi$) sites, which makes it hard to distinguish the best site with highest  chance of good science returns. 

\subsubsection*{Results}
\begin{figure}[t]
    \centering
    \includegraphics[width=0.36\textwidth]{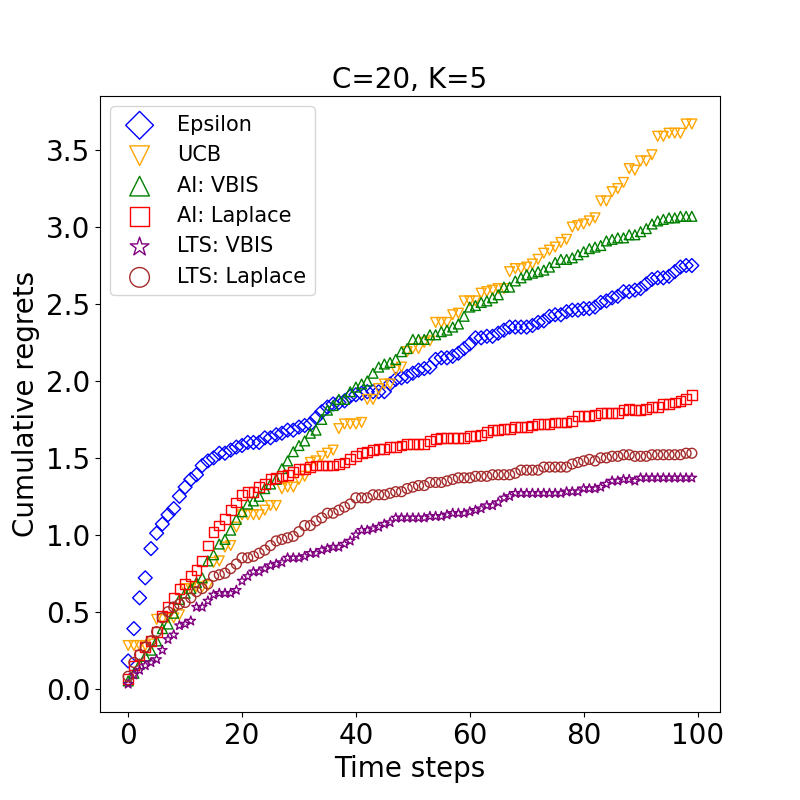}
    \caption{Cumulative regrets among different approaches in case of $C$=$20$ and $K$=$5$. Prior preference is non-reward seeking and favors more uncertain outcome distributions, i.e. $p_{ev}(o) = [0.4, 0.6]^T$.}
    \label{fig: ev_04_06}
    \vspace{-0.15in}
\end{figure}

The cumulative regret (\ref{eq: regret}) was used to compare the performance of all approaches. 
Fig. \ref{fig: comparison_cumulative_regrets} shows the characteristic results, including those obtained when the easiest (bottom right; the most amount of information and the fewest number of options) and most difficult (top right; lowest amount of information and most options) conditions are used. As can be seen from this figure, as the difficulty of the CMAB problem increases (counterclockwise from the bottom right to top right), the values of cumulative regrets tend to increase regardless of which action selection strategy is used, nevertheless the both AI VBIS and AI Laplace robots greatly outperforms the others (Note: the slight performance differences between the AI VBIS and AI Laplace robots are mainly from the accuracy of the posterior distribution approximation). This is owing to the fact that, as shown in (\ref{eq: efe_each_option}), AI can rigorously evaluate the uncertainty of each option and quickly find large $\psi$ options, 
and prefers those options under the influence of the prior preference $p_{ev}(o)$. On the other hand, when the prior preference is non-reward seeking and favors more uncertain outcome distributions (i.e. $p_{ev}(o) = [0.4, 0.6]^T$), the AI robots perform as comparable as other action selection strategies as presented in Fig. \ref{fig: ev_04_06}. In such a case, in order to assess the performance of the AI robots, other evaluation metric such as the Kullback-Leibler divergence may be utilized (i.e. the extent to which the outcome distribution obtained by executing the option preferred by the AI robots are aligned with the prior preference).
In terms of computation cost, although the cost of AI VBIS approach is higher than other strategies due to importance sampling, this can be significantly reduced by computing sample weights in parallel. With regard to AI Laplace, while the average single action selection time ($0.010$ sec in case of $K=10$, $C=10$) is again higher than that of $\epsilon$-greedy ($3.42$E-06 sec in the same condition), still the computation cost may not be problematic since the number of total iterations is small enough (at most $10^2$) and the performance of the AI Laplace robot outperforms the baseline methods. Even in the most difficult case ($K=40, C=20$), AI Laplace takes only $0.26$ sec on average, compared to $5.48$E-06 sec for $\epsilon$-greedy.

%% file: conclusion.tex
In this paper, we applied active inference as an action selection strategy for contextual multi-armed bandit (CMAB) problems. We showed how to derive the expected free energy (EFE) in categorical outcome CMAB problems and introduced variational and Laplace approximation algorithms to cope with analytically intractable EFE terms. 
The proposed active inference 
strategy was demonstrated in a simulation for an autonomous remote science site selection scenario, and its cumulative regret was compared with the ones of other strategies. It was found that when the prior preference is reward-seeking, active inference brings significant advantages, due to its ability to evaluate option uncertainty and bias towards preferred outcomes. 

The primary objective of this study was to validate the feasibility and performance of active inference strategy in practical CMAB problems using proof of concept simulations that reflect key aspects of the motivating remote space science exploration problem. Future work will confirm the effectiveness of this strategy using OceanWATERS, 
\cite{oceanwaters}, 
a testbed for the Europa lander recently released by NASA. In this work, the value of prior preferences was determined and fixed {\it a priori}, although adjusting this value according to search context, the regret analysis of active inference action selection strategies and results obtained from the actual action execution are also future tasks. 